\documentclass[letter]{article} % For LaTeX2e
\usepackage{iclr2018_conference,times}
\usepackage[hidelinks=true]{hyperref}
\usepackage{url}

\usepackage{latexsym}

\usepackage{enumitem}
\usepackage{xspace}
\usepackage{booktabs}
\usepackage{multirow}
\usepackage{tikz}
\usepackage[font=small]{caption}
\usepackage{subcaption}

\usepackage{amsmath}
\usepackage{amsfonts} % mathbb
\usepackage{amssymb} % \mathbb and extra symbols

\newcommand{\ptb}{Penn Treebank\xspace}
\newcommand{\wikitexttwo}{Wikitext-2\xspace}
\newcommand{\enwik}{Enwik8\xspace}
\newcommand{\nlltoppl}[1]{%
  \pgfmathparse{exp(#1)}%
  \pgfmathprintnumber[fixed,zerofill,precision=1]{\pgfmathresult}}
\newcommand{\nlltobpc}[1]{%
  \pgfmathparse{log2(exp(#1))}%
  \pgfmathprintnumber[fixed,zerofill,precision=2]{\pgfmathresult}}

\title{On the State of the Art of Evaluation in\\ Neural Language Models}

\author{G\'abor Melis$^\dag$, Chris Dyer$^\dag$, Phil Blunsom$^{\dag\ddag}$ \\
  {\tt \{melisgl,cdyer,pblunsom\}@google.com}\\
  $^\dag$DeepMind\\
  $^\ddag$University of Oxford
}

\iclrfinalcopy % Uncomment for camera-ready version, but NOT for submission.

\begin{document}

\maketitle

\begin{abstract}
  Ongoing innovations in recurrent neural network
  architectures have provided a steady influx of apparently
  state-of-the-art results on language modelling benchmarks. However,
  these have been evaluated using differing codebases and limited
  computational resources, which represent uncontrolled sources of
  experimental variation. We reevaluate several popular
  architectures and regularisation methods with large-scale automatic
  black-box hyperparameter tuning and arrive at the somewhat
  surprising conclusion that standard LSTM architectures, when
  properly regularised, outperform more recent models. We establish a
  new state of the art on the Penn Treebank and Wikitext-2 corpora, as
  well as strong baselines on the Hutter Prize dataset.
\end{abstract}

\section{Introduction}

The scientific process by which the deep learning research community
operates is guided by empirical studies that evaluate the relative
quality of models. Complicating matters, the measured performance of a
model depends not only on its architecture (and data), but it can
strongly depend on hyperparameter values that affect learning,
regularisation, and capacity. This hyperparameter dependence is an
often inadequately controlled source of variation in experiments,
which creates a risk that empirically unsound claims will be reported.

In this paper, we use a black-box hyperparameter optimisation
technique to control for hyperparameter effects while comparing the
relative performance of language modelling architectures based on
LSTMs, Recurrent Highway Networks
\citep{DBLP:journals/corr/ZillySKS16} and NAS \citep{zoph2016neural}.
We specify flexible, parameterised model families with the ability to
adjust embedding and recurrent cell sizes for a given parameter budget
and with fine grain control over regularisation and learning
hyperparameters.

Once hyperparameters have been properly controlled for, we find that
LSTMs outperform the more recent models, contra the published claims.
Our result is therefore a demonstration that replication failures can
happen due to poorly controlled hyperparameter variation, and this
paper joins other recent papers in warning of the under-acknowledged
existence of replication failure in deep
learning~\citep{henderson2017deep,DBLP:journals/corr/ReimersG17a}.
However, we do show that careful controls are possible, albeit at
considerable computational cost.

Several remarks can be made in light of these results. First, as
(conditional) language models serve as the central building block of
many tasks, including machine translation, there is little reason to
expect that the problem of unreliable evaluation is unique to the
tasks discussed here. However, in machine translation, carefully
controlling for hyperparameter effects would be substantially more
expensive because standard datasets are much larger. Second, the
research community should strive for more consensus about appropriate
experimental methodology that balances costs of careful
experimentation with the risks associated with false claims. Finally, more attention should be paid to
hyperparameter sensitivity. Models that introduce many new
hyperparameters or which perform well only in narrow ranges of
hyperparameter settings should be identified as such as part of
standard publication practice.

\section{Models}

\begin{figure}[!t]\centering
  \captionsetup[subfigure]{justification=centering}
  \begin{subfigure}{0.49\linewidth}
    \includegraphics[width=1.0\linewidth,trim={1.0cm 0 1.0cm 0}]
                    {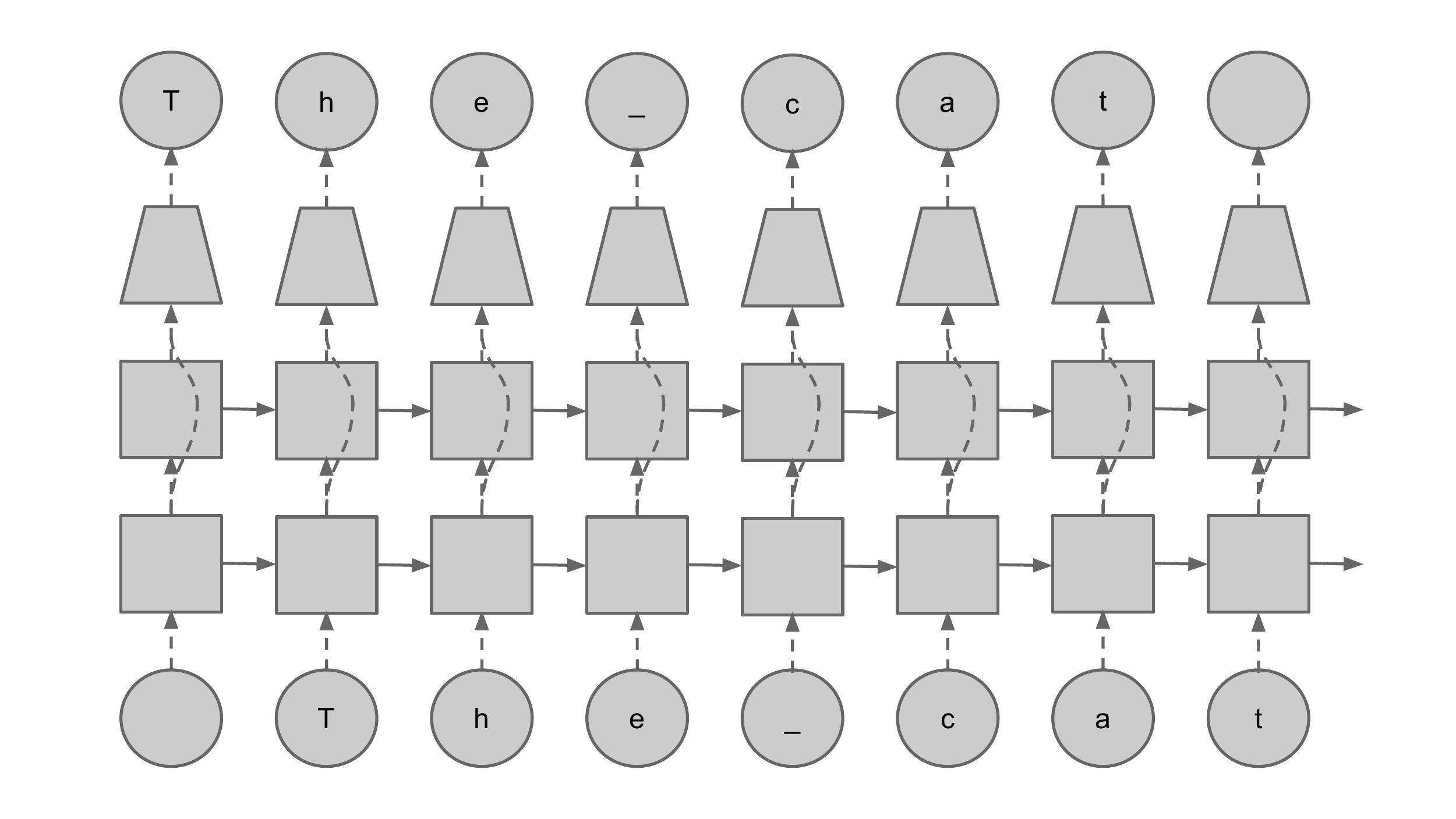}
    \caption{two-layer LSTM/NAS with skip connections}
    \label{fig:lstm}
  \end{subfigure}
  \hfill
  \begin{subfigure}{0.49\linewidth}
    \includegraphics[width=1.0\linewidth,trim={1.0cm 0 1.0cm 0}]
                    {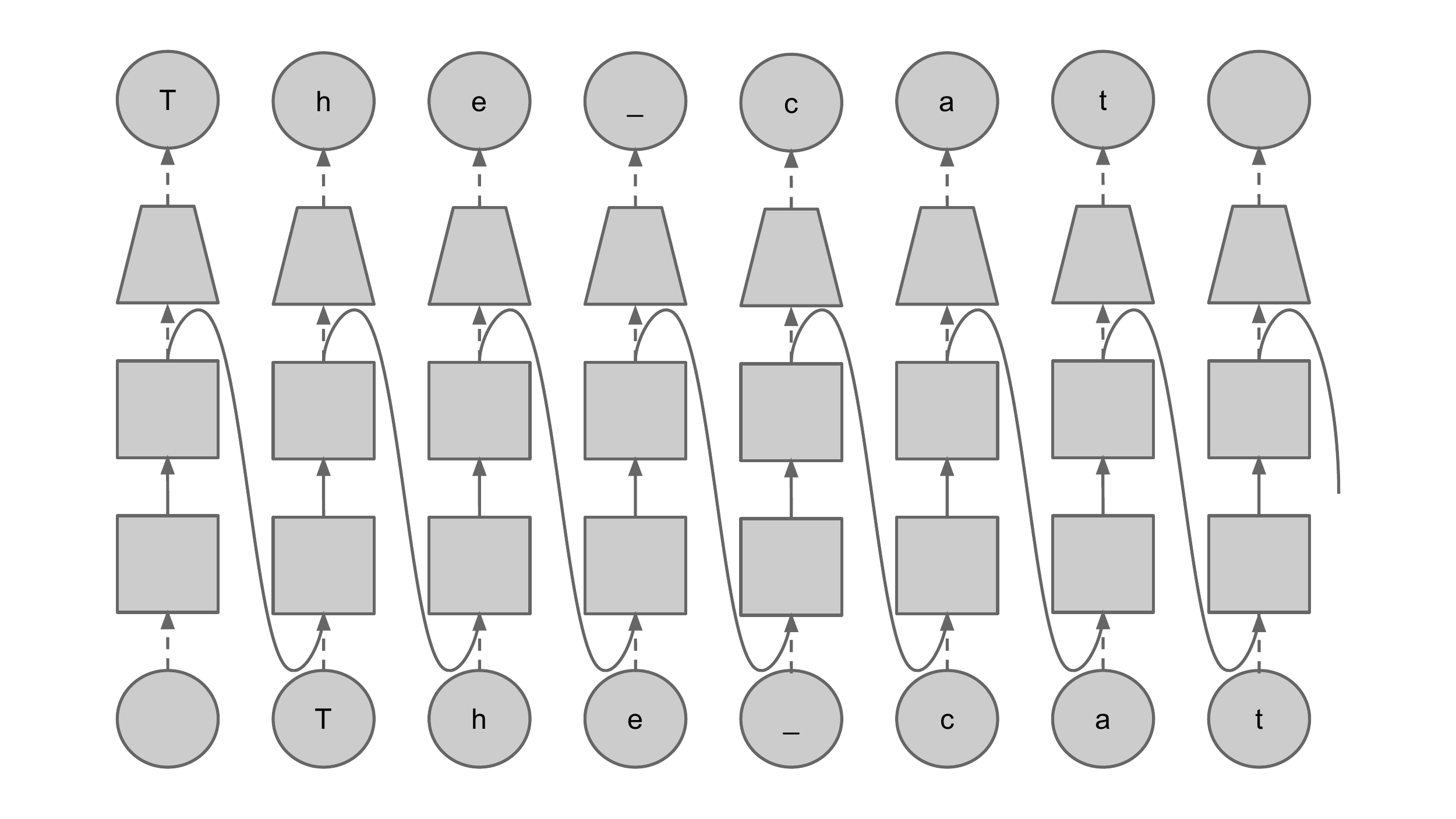}
    \caption{RHN with two processing steps per input}
    \label{fig:rhn}
  \end{subfigure}
  \caption{\small Recurrent networks with optional down-projection,
    per-step and per-sequence dropout (dashed and solid lines).}
\end{figure}

Our focus is on three recurrent architectures:
\begin{itemize}
\item The Long Short-Term Memory \citep{hochreiter:1997:lstm} serves
  as a well known and frequently used baseline.
\item The recently proposed Recurrent Highway Network
  \citep{DBLP:journals/corr/ZillySKS16} is chosen because it has
  demonstrated state-of-the-art performance on a number of datasets.
\item Finally, we also include NAS \citep{zoph2016neural}, because of
  its impressive performance and because its architecture was the
  result of an automated reinforcement learning based optimisation
  process.
\end{itemize}
Our aim is strictly to do better model comparisons for these
architectures and we thus refrain from including techniques that are
known to push perplexities even lower, but which are believed to be
largely orthogonal to the question of the relative merits of these
recurrent cells. In parallel work with a remarkable overlap with ours,
\citet{DBLP:journals/corr/abs-1708-02182} demonstrate the utility of
adding a Neural Cache \citep{DBLP:journals/corr/GraveJU16}. Building
on their work, \citet{krause2017dynamic} show that Dynamic Evaluation
\citep{DBLP:journals/corr/Graves13} contributes similarly to the final
perplexity.

As pictured in Fig.~\ref{fig:lstm}, our models with LSTM or NAS cells
have all the standard components: an input embedding lookup table,
recurrent cells stacked as layers with additive skip connections
combining outputs of all layers to ease optimisation. There is an
optional down-projection whose presence is governed by a
hyperparameter from this combined output to a smaller space which
reduces the number of output embedding parameters. Unless otherwise
noted, input and output embeddings are shared, see
\citep{DBLP:journals/corr/InanKS16} and
\citep{DBLP:journals/corr/PressW16}.

Dropout is applied to feedforward connections denoted by dashed arrows
in the figure. From the bottom up: to embedded inputs (\textit{input
  dropout}), to connections between layers (\textit{intra-layer
  dropout}), to the combined and the down-projected outputs
(\textit{output dropout}). All these dropouts have random masks drawn
independently per time step, in contrast to the dropout on recurrent
states where the same mask is used for all time steps in the sequence.

RHN based models are typically conceived of as a single horizontal ``highway''
to emphasise how the recurrent state is processed through time. In
Fig. \ref{fig:rhn}, we choose to draw their schema in a way that makes
the differences from LSTMs immediately apparent. In a nutshell, the
RHN state is passed from the topmost layer to the lowest layer of the
next time step. In contrast, each LSTM layer has its own recurrent
connection and state.

The same dropout variants are applied to all three model types, with
the exception of intra-layer dropout which does not apply to RHNs
since only the recurrent state is passed between the layers. For the
recurrent states, all architectures use either variational dropout
\citep[\emph{state dropout}]{gal2016theoretically}\footnote{Of the two parameterisations, we used the one in
  which there is further sharing of masks between gates rather than
  independent noise for the gates.} or recurrent dropout
\citep{DBLP:journals/corr/SemeniutaSB16}, unless explicitly noted otherwise.

\section{Experimental Setup}

\subsection{Datasets}

We compare models on three datasets. The smallest of them is the \ptb
corpus by \citet{marcus1993building} with preprocessing from
\citet{mikolov2010recurrent}. We also include another word level
corpus: \wikitexttwo by \citet{DBLP:journals/corr/MerityXBS16}. It is
about twice the size of \ptb with a larger vocabulary and much lighter
preprocessing. The third corpus is \enwik from the Hutter Prize
dataset \citep{hutter2012}. Following common practice, we use the first
90 million characters for training, and the remaining 10 million
evenly split between validation and test.

\section{Training details}

When training word level models we follow common practice and use a
batch size of 64, truncated backpropagation with 35 time steps, and we
feed the final states from the previous batch as the initial state of
the subsequent one. At the beginning of training and test time, the
model starts with a zero state. To bias the model towards being able
to easily start from such a state at test time, during training, with
probability 0.01 a constant zero state is provided as the initial
state.

Optimisation is performed by Adam \citep{kingma2014adam} with
$\beta_1=0$ but otherwise default parameters ($\beta_2=0.999$,
$\epsilon=10^{-9}$). Setting $\beta_1$ so turns off the exponential
moving average for the estimates of the means of the gradients and
brings Adam very close to RMSProp without momentum, but due to Adam's
bias correction, larger learning rates can be used.

Batch size is set to 64. The learning rate is multiplied by 0.1
whenever validation performance does not improve ever during 30
consecutive checkpoints. These checkpoints are performed after every
100 and 200 optimization steps for \ptb and \wikitexttwo,
respectively.

For character level models (i.e. \enwik), the differences are:
truncated backpropagation is performed with 50 time steps. Adam's
parameters are $\beta_2=0.99$, $\epsilon=10^{-5}$. Batch size is 128.
Checkpoints are only every 400 optimisation steps and embeddings are
not shared.

\begin{table}[t]
  \small
  \centering
  \begin{tabular}{@{}lrrrr@{}}
    \toprule
    Model & Size & Depth & Valid & Test \\
    \midrule
    Medium LSTM, \citet{DBLP:journals/corr/ZarembaSV14}    & 10M &  2 &  86.2 &  82.7 \\
    Large LSTM, \citet{DBLP:journals/corr/ZarembaSV14}     & 24M &  2 &  82.2 &  78.4 \\
    VD LSTM, \citet{DBLP:journals/corr/PressW16}           & 51M &  2 &  75.8 &  73.2 \\
    VD LSTM, \citet{DBLP:journals/corr/InanKS16}           &  9M &  2 &  77.1 &  73.9 \\
    VD LSTM, \citet{DBLP:journals/corr/InanKS16}           & 28M &  2 &  72.5 &  69.0 \\
    VD RHN, \citet{DBLP:journals/corr/ZillySKS16}          & 24M & 10 &  67.9 &  65.4 \\
    NAS, \citet{zoph2016neural}                            & 25M &  - &  -    &  64.0 \\
    NAS, \citet{zoph2016neural}                            & 54M &  - &  -    &  62.4 \\
    AWD-LSTM, \citet{DBLP:journals/corr/abs-1708-02182} \textdagger    & 24M &  3 &  60.0 &  57.3 \\
    \midrule
    \midrule
    LSTM & \multirow{5}{*}{10M} & 1 & \nlltoppl{4.124} & \nlltoppl{4.088} \\
    LSTM &                      & 2 & \nlltoppl{4.143} & \nlltoppl{4.107} \\
    LSTM &                      & 4 & \nlltoppl{4.134} & \nlltoppl{4.096} \\
    RHN  &                      & 5 & \nlltoppl{4.189} & \nlltoppl{4.151} \\
    NAS  &                      & 1 & \nlltoppl{4.184} & \nlltoppl{4.139} \\
    \midrule
    LSTM & \multirow{5}{*}{24M} & 1 & \nlltoppl{4.117} & \nlltoppl{4.086} \\
    LSTM &                      & 2 & \nlltoppl{4.128} & \nlltoppl{4.087} \\
    LSTM &                      & 4 & \nlltoppl{4.110} & \nlltoppl{4.065} \\
    RHN  &                      & 5 & \nlltoppl{4.172} & \nlltoppl{4.131} \\
    NAS  &                      & 1 & \nlltoppl{4.129} & \nlltoppl{4.090} \\
    \bottomrule
  \end{tabular}
  \caption{\small Validation and test set perplexities on \ptb for
    models with different numbers of parameters and depths. All
    results except those from Zaremba are with shared input and output
    embeddings. VD stands for Variational Dropout from
    \citet{gal2016theoretically}. \textdagger: parallel work.}
  \label{tab:ptb-results}
\end{table}

\section{Evaluation}

For evaluation, the checkpoint with the best validation perplexity
found by the tuner is loaded and the model is applied to the test set
with a batch size of 1. For the word based datasets, using the
training batch size makes results worse by 0.3 PPL while \enwik is
practically unaffected due to its evaluation and training sets being
much larger. Preliminary experiments indicate that MC averaging would
bring a small improvement of about 0.4 in perplexity and 0.005 in bits
per character, similar to the results of \citet{gal2016theoretically},
while being a 1000 times more expensive which is prohibitive on larger
datasets. Therefore, throughout we use the mean-field approximation
for dropout at test time.

\subsection{Hyperparameter Tuning}

Hyperparameters are optimised by Google Vizier
\citep{golovin2017google}, a black-box hyperparameter tuner based on
batched GP bandits using the expected improvement acquisition function
\citep{JMLR:v15:desautels14a}. Tuners of this nature are generally
more efficient than grid search when the number of hyperparameters is
small. To keep the problem tractable, we restrict the set of
hyperparameters to \textit{learning rate}, \textit{input embedding
  ratio}, \textit{input dropout}, \textit{state dropout},
\textit{output dropout}, \textit{weight decay}. For deep LSTMs, there
is an extra hyperparameter to tune: \textit{intra-layer dropout}. Even
with this small set, thousands of evaluations are required to reach
convergence.

\paragraph{Parameter budget.} Motivated by recent results from
\citet{collins2016capacity}, we compare models on the basis of the
total number of trainable parameters as opposed to the number of
hidden units. The tuner is given control over the presence and size of
the down-projection, and thus over the tradeoff between the number of
embedding vs. recurrent cell parameters. Consequently, the cells'
hidden size and the embedding size is determined by the actual
parameter budget, depth and the \textit{input embedding ratio}
hyperparameter.

For \enwik there are relatively few parameters in the embeddings since
the vocabulary size is only 205. Here we choose not to share
embeddings and to omit the down-projection unconditionally.

\section{Results}

\subsection{Penn Treebank}

We tested LSTMs of various depths and an RHN of depth 5 with parameter
budgets of 10 and 24 million matching the sizes of the Medium and
Large LSTMs by \citep{DBLP:journals/corr/ZarembaSV14}. The results are
summarised in Table \ref{tab:ptb-results}.

Notably, in our experiments even the RHN with only 10M parameters has
better perplexity than the 24M one in the original publication. Our
24M version improves on that further. However, a shallow LSTM-based
model with only 10M parameters enjoys a very comfortable margin over
that, with deeper models following near the estimated noise range. At
24M, all depths obtain very similar results, reaching \nlltoppl{4.065}
at depth 4. Unsurprisingly, NAS whose architecture was chosen based on
its performance on this dataset does almost equally well, even better
than in \citet{zoph2016neural}.

\subsection{Wikitext-2}

\begin{table}[t]
  \small
  \centering
  \begin{tabular}{@{}lrrrr@{}}
    \toprule
    Model & Size & Depth & Valid & Test \\
    \midrule
    VD LSTM, \citet{DBLP:journals/corr/MerityXBS16}           & 20M &  2 & 101.7 &  96.3 \\
    VD+Zoneout LSTM, \citet{DBLP:journals/corr/MerityXBS16}   & 20M &  2 & 108.7 & 100.9 \\
    VD LSTM, \citet{DBLP:journals/corr/InanKS16}              & 22M &  2 &  91.5 &  87.7 \\
    AWD-LSTM, \citet{DBLP:journals/corr/abs-1708-02182} \textdagger & 33M &  3 &  68.6 &  65.8 \\
    \midrule
    \midrule
    LSTM (tuned for PTB) & \multirow{5}{*}{10M} & 1 & \nlltoppl{4.482} & \nlltoppl{4.421} \\
    LSTM                 &                      & 1 & \nlltoppl{4.286} & \nlltoppl{4.235} \\
    LSTM                 &                      & 2 & \nlltoppl{4.301} & \nlltoppl{4.259} \\
    LSTM                 &                      & 4 & \nlltoppl{4.361} & \nlltoppl{4.308} \\
    RHN                  &                      & 5 & \nlltoppl{4.425} & \nlltoppl{4.376} \\
    NAS                  &                      & 1 & \nlltoppl{4.377} & \nlltoppl{4.329} \\
    \midrule
    LSTM (tuned for PTB) & \multirow{5}{*}{24M} & 1 & \nlltoppl{4.380} & \nlltoppl{4.335} \\
    LSTM                 &                      & 1 & \nlltoppl{4.238} & \nlltoppl{4.188} \\
    LSTM                 &                      & 2 & \nlltoppl{4.236} & \nlltoppl{4.188} \\
    LSTM                 &                      & 4 & \nlltoppl{4.256} & \nlltoppl{4.213} \\
    RHN                  &                      & 5 & \nlltoppl{4.358} & \nlltoppl{4.325} \\
    NAS                  &                      & 1 & \nlltoppl{4.290} & \nlltoppl{4.246} \\
    \bottomrule
  \end{tabular}
  \caption{\small Validation and test set perplexities on
    \wikitexttwo. All results are with shared input and output
    embeddings. \textdagger: parallel work.}
  \label{tab:wikitext2-results}
\end{table}

\wikitexttwo is not much larger than \ptb, so it is not surprising
that even models tuned for \ptb perform reasonably on this dataset,
and this is in fact how results in previous works were produced. For a
fairer comparison, we also tune hyperparameters on the same dataset.
In Table \ref{tab:wikitext2-results}, we report numbers for both
approaches. All our results are well below the previous state of the
are for models without dynamic evaluation or caching. That said, our
best result, \nlltoppl{4.188} compares favourably even to the Neural
Cache \citep{DBLP:journals/corr/GraveJU16} whose innovations are
fairly orthogonal to the base model.

Shallow LSTMs do especially well here. Deeper models have gradually
degrading perplexity, with RHNs lagging all of them by a significant
margin. NAS is not quite up there with the LSTM suggesting its
architecture might have overfitted to \ptb, but data for deeper
variants would be necessary to draw this conclusion.

\subsection{Enwik8}

In contrast to the previous datasets, our numbers on this task
(reported in BPC, following convetion) are slightly off the state of
the art. This is most likely due to optimisation being limited to 14
epochs which is about a tenth of what the model of
\citet{DBLP:journals/corr/ZillySKS16} was trained for. Nevertheless,
we match their smaller RHN with our models which are very close to
each other. NAS lags the other models by a surprising margin at this
task.

\section{Analysis}

On two of the three datasets, we improved previous results
substantially by careful model specification and hyperparameter
optimisation, but the improvement for RHNs is much smaller compared to
that for LSTMs. While it cannot be ruled out that our particular setup
somehow favours LSTMs, we believe it is more likely that this effect
arises due to the original RHN experimental condition having been
tuned more extensively (this is nearly unavoidable during model
development).

Naturally, NAS benefitted only to a limited degree from our tuning,
since the numbers of \citet{zoph2016neural} were already produced by
employing similar regularisation methods and a grid search. The small
edge can be attributed to the suboptimality of grid search (see
Section~\ref{sec:sensitivity}).

In summary, the three recurrent cell architectures are closely matched
on all three datasets, with minuscule differences on \enwik where
regularisation matters the least. These results support the claims of
\citet{collins2016capacity}, that capacities of various cells are very
similar and their apparent differences result from trainability and
regularisation. While comparing three similar architectures cannot
prove this point, the inclusion of NAS certainly gives it more
credence. This way we have two of the best human designed and one
machine optimised cell that was the top performer among thousands of
candidates.

\begin{table}[t]
  \small
  \centering
  \begin{tabular}{@{}lrrrr@{}}
    \toprule
    Model & Size & Depth & Valid & Test \\
    \midrule
    Stacked LSTM, \citet{DBLP:journals/corr/Graves13}            & 21M &  7 & - & 1.67 \\
    Grid LSTM, \citet{DBLP:journals/corr/KalchbrennerDG15}& 17M  &  6 & - & 1.47 \\
    MI-LSTM, \citet{DBLP:journals/corr/WuZZBS16}                 & 17M &  1 & - & 1.44 \\
    LN HM-LSTM, \citet{DBLP:journals/corr/ChungAB16}             & 35M &  3 & - & 1.32 \\
    ByteNet, \citet{DBLP:journals/corr/KalchbrennerESO16}        &   - & 25 & - & 1.31 \\
    VD RHN, \citet{DBLP:journals/corr/ZillySKS16}                & 23M &  5 & - & 1.31 \\
    VD RHN, \citet{DBLP:journals/corr/ZillySKS16}                & 21M & 10 & - & 1.30 \\
    VD RHN, \citet{DBLP:journals/corr/ZillySKS16}                & 46M & 10 & - & 1.27 \\
    \midrule
    \midrule
    LSTM & \multirow{3}{*}{27M} & 4 & \nlltobpc{0.897} & \nlltobpc{0.908} \\
    RHN  &                      & 5 & \nlltobpc{0.900} & \nlltobpc{0.908} \\
    NAS  &                      & 4 & \nlltobpc{0.954} & \nlltobpc{0.969} \\
    \midrule
    LSTM & \multirow{2}{*}{46M} & 4 & \nlltobpc{0.886} & \nlltobpc{0.898} \\
    RHN  &                      & 5 & \nlltobpc{0.892} & \nlltobpc{0.898} \\
    NAS  &                      & 4 & \nlltobpc{0.917} & \nlltobpc{0.925} \\
    \bottomrule
  \end{tabular}
  \caption{\small Validation and test set BPCs on \enwik from the
    Hutter Prize dataset.}
  \label{tab:enwik8-wt-results}
\end{table}

\subsection{The Effect of Individual Features}

Down-projection was found to be very beneficial by the tuner for some
depth/budget combinations. On \ptb, it improved results by about 2--5
perplexity points at depths 1 and 2 at 10M, and depth 1 at 24M,
possibly by equipping the recurrent cells with more capacity. The very
same models benefited from down-projection on \wikitexttwo, but even
more so with gaps of about 10--18 points which is readily explained by
the larger vocabulary size.

We further measured the contribution of other features of the models
in a series of experiments. See Table \ref{tab:ptb-variant-results}.
To limit the number of resource used, in these experiments only
individual features were evaluated (not their combinations) on \ptb at
the best depth for each architecture (LSTM or RHN) and parameter
budget (10M or 24M) as determined above.

First, we untied input and output embeddings which made perplexities
worse by about 6 points across the board which is consistent with the
results of \citet{DBLP:journals/corr/InanKS16}.

Second, without variational dropout the RHN models suffer quite a bit
since there remains no dropout at all in between the layers. The deep
LSTM also sees a similar loss of perplexity as having intra-layer
dropout does not in itself provide enough regularisation.

Third, we were also interested in how recurrent dropout
\citep{DBLP:journals/corr/SemeniutaSB16} would perform in lieu of
variational dropout. Dropout masks were shared between time steps in
both methods, and our results indicate no consistent advantage to
either of them.

\subsection{Model Selection}

With a large number of hyperparameter combinations evaluated, the
question of how much the tuner overfits arises. There are multiple
sources of noise in play,
\begin{enumerate}[label=(\alph*),nolistsep]
\item \label{it:badfp}non-deterministic ordering of floating-point
  operations in optimised linear algebra routines,
\item \label{it:seeds}different initialisation seeds,
\item \label{it:finite}the validation and test sets being finite
  samples from a infinite population.
\end{enumerate}
To assess the severity of these issues, we conducted the following
experiment: models with the best hyperparameter settings for \ptb and
\wikitexttwo were retrained from scratch with various initialisation
seeds and the validation and test scores were recorded. If during
tuning, a model just got a lucky run due to a combination of
\ref{it:badfp} and \ref{it:seeds}, then retraining with the same
hyperparameters but with different seeds would fail to reproduce the
same good results.

There are a few notable things about the results. First, in our
environment (Tensorflow with a single GPU) even with the same seed as
the one used by the tuner, the effect of \ref{it:badfp} is almost as
large as that of \ref{it:badfp} and \ref{it:seeds} combined. Second,
the variance induced by \ref{it:badfp} and \ref{it:seeds} together is
roughly equivalent to an absolute difference of 0.4 in perplexity on
\ptb and 0.5 on \wikitexttwo. Third, the validation perplexities of
the best checkpoints are about one standard deviation lower than the
sample mean of the reruns, so the tuner could fit the noise only to a
limited degree.

Because we treat our corpora as a single sequence, test set contents
are not i.i.d., and we cannot apply techniques such as the bootstrap
to assess \ref{it:finite}. Instead, we looked at the gap between
validation and test scores as a proxy and observed that it is very
stable, contributing variance of 0.12--0.3 perplexity to the final
results on \ptb and \wikitexttwo, respectively.

We have not explicitly dealt with the unknown uncertainty remaining in
the Gaussian Process that may affect model comparisons, apart from
running it until apparent convergence. All in all, our findings
suggest that a gap in perplexity of 1.0 is a statistically robust
difference between models trained in this way on these datasets. The
distribution of results was approximately normal with roughly the same
variance for all models, so we still report numbers in a tabular form
instead of plotting the distribution of results, for example in a
violin plot \citep{hintze1998violin}.

\begin{table}[t]
  \small
  \centering
  \begin{tabular}{@{}l crr crr@{}}
    \toprule
          & \multicolumn{3}{c}{Size 10M} & \multicolumn{3}{c}{Size 24M} \\
            \cmidrule(l){2-4}              \cmidrule(l){5-7}
    Model & Depth & Valid & Test         & Depth & Valid & Test \\
    \midrule
    LSTM                  & 1 & \nlltoppl{4.124} & \nlltoppl{4.088} & 4 & \nlltoppl{4.110} & \nlltoppl{4.065} \\
    \midrule
    \midrule
    - Shared Embeddings   & 1 & \nlltoppl{4.214} & \nlltoppl{4.177} & 4 & \nlltoppl{4.184} & \nlltoppl{4.147} \\
    - Variational Dropout & 1 & \nlltoppl{4.142} & \nlltoppl{4.114} & 4 & \nlltoppl{4.194} & \nlltoppl{4.166} \\
    + Recurrent Dropout   & 1 & \nlltoppl{4.140} & \nlltoppl{4.105} & 4 & \nlltoppl{4.178} & \nlltoppl{4.142} \\
    + Untied gates        & 1 & \nlltoppl{4.117} & \nlltoppl{4.076} & 4 & \nlltoppl{4.159} & \nlltoppl{4.116} \\
    + Tied gates          & 1 & \nlltoppl{4.122} & \nlltoppl{4.088} & 4 & \nlltoppl{4.101} & \nlltoppl{4.060} \\
    \midrule
    RHN                   & 5 & \nlltoppl{4.189} & \nlltoppl{4.151} & 5 & \nlltoppl{4.172} & \nlltoppl{4.131} \\
    \midrule
    \midrule
    - Shared Embeddings   & 5 & \nlltoppl{4.281} & \nlltoppl{4.242} & 5 & \nlltoppl{4.210} & \nlltoppl{4.168} \\
    - Variational Dropout & 5 & \nlltoppl{4.309} & \nlltoppl{4.272} & 5 & \nlltoppl{4.314} & \nlltoppl{4.273} \\
    + Recurrent Dropout   & 5 & \nlltoppl{4.182} & \nlltoppl{4.143} & 5 & \nlltoppl{4.149} & \nlltoppl{4.111} \\
    \bottomrule
  \end{tabular}
  \caption{\small Validation and test set perplexities on \ptb for
    variants of our best LSTM and RHN models of two sizes.}
  \label{tab:ptb-variant-results}
\end{table}

\subsection{Sensitivity}
\label{sec:sensitivity}

\begin{figure*}[!t]\centering
  \includegraphics[width=\linewidth,trim={0 0 0 0}]
                  {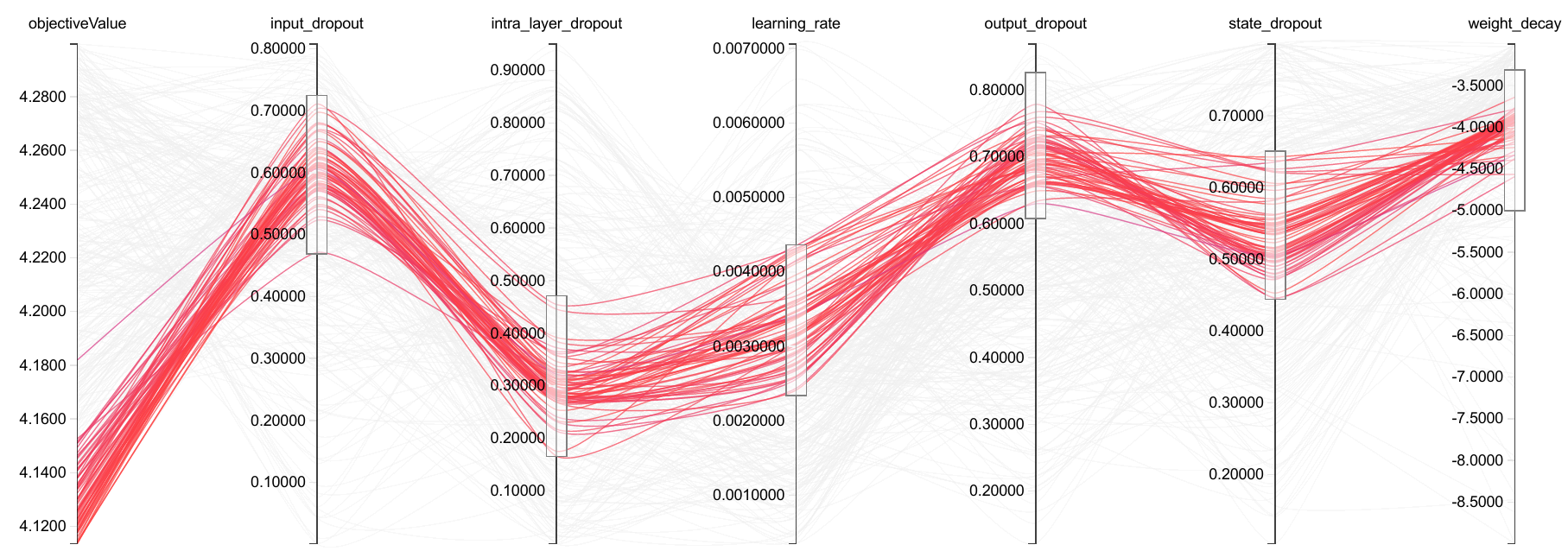}
  \caption{Average per-word negative log-likelihoods of
    hyperparameter combinations in the neighbourhood of the best
    solution for a 4-layer LSTM with 24M weights on the \ptb dataset.}
  \label{fig:hp-sensitivity}
\end{figure*}

To further verify that the best hyperparameter setting found by the
tuner is not a fluke, we plotted the validation loss against the
hyperparameter settings. Fig. \ref{fig:hp-sensitivity} shows one such
typical plot, for a 4-layer LSTM. We manually restricted the ranges
around the best hyperparameter values to around 15--25\% of the entire
tuneable range, and observed that the vast majority of settings in
that neighbourhood produced perplexities within 3.0 of the best value.
Widening the ranges further leads to quickly deteriorating results.

Satisfied that the hyperparameter surface is well behaved, we
considered whether the same results could have possibly been achieved
with a simple grid search. Omitting \textit{input embedding ratio}
because the tuner found having a down-projection suboptimal almost
non-conditionally for this model, there remain six hyperparameters to
tune. If there were 5 possible values on the grid for each
hyperparameter (with one value in every 20\% interval), then we would
need $6^5$, nearly 8000 trials to get within 3.0 of the best
perplexity achieved by the tuner in about 1500 trials.

\subsection{Tying LSTM gates}

Normally, LSTMs have two independent gates controlling the retention
of cell state and the admission of updates (Eq. \ref{eq:lstm-untied}).
A minor variant which reduces the number of parameters at the loss of
some flexibility is to tie the input and forget gates as in Eq.
\ref{eq:lstm-tied}. A possible middle ground that keeps the number of
parameters the same but ensures that values of the cell state $c$
remain in $[-1, 1]$ is to cap the input gate as in Eq.
\ref{eq:lstm-capped}.
\begin{align}
  \label{eq:lstm-untied} \mathbf{c}_t &= \mathbf{f}_t \odot \mathbf{c}_{t-1} + \mathbf{i}_t \odot \mathbf{j}_t \\
  \label{eq:lstm-tied}   \mathbf{c}_t &= \mathbf{f}_t \odot \mathbf{c}_{t-1} + (1-\mathbf{f}_t) \odot \mathbf{j}_t \\
  \label{eq:lstm-capped} \mathbf{c}_t &= \mathbf{f}_t \odot \mathbf{c}_{t-1} + \min(1-\mathbf{f}_t, \mathbf{i}_t) \odot \mathbf{j}_t
\end{align}
Where the equations are based on the formulation of
\citet{DBLP:journals/corr/SakSB14}. All LSTM models in this paper use
the third variant, except those titled ``Untied gates'' and ``Tied
gates'' in Table \ref{tab:ptb-variant-results} corresponding to Eq.
\ref{eq:lstm-untied} and \ref{eq:lstm-tied}, respectively.

The results show that LSTMs are insensitive to these changes and the
results vary only slightly even though more hidden units are allocated
to the tied version to fill its parameter budget. Finally, the numbers
suggest that deep LSTMs benefit from bounded cell states.

\section{Conclusion}

During the transitional period when deep neural language models began
to supplant their shallower predecessors, effect sizes tended to be
large, and robust conclusions about the value of the modelling
innovations could be made, even in the presence of poorly controlled
``hyperparameter noise.'' However, now that the neural revolution is
in full swing, researchers must often compare competing deep
architectures. In this regime, effect sizes tend to be much smaller,
and more methodological care is required to produce reliable results.
Furthermore, with so much work carried out in parallel by a growing
research community, the costs of faulty conclusions are increased.

Although we can draw attention to this problem, this paper does not
offer a practical methodological solution beyond establishing reliable
baselines that can be the benchmarks for subsequent work. Still, we
demonstrate how, with a huge amount of computation, noise levels of
various origins can be carefully estimated and models meaningfully
compared. This apparent tradeoff between the amount of computation and
the reliability of results seems to lie at the heart of the matter.
Solutions to the methodological challenges must therefore make model
evaluation cheaper by, for instance, reducing the number of
hyperparameters and the sensitivity of models to them, employing
better hyperparameter optimisation strategies, or by defining
``leagues'' with predefined computational budgets for a single model
representing different points on the tradeoff curve.

\bibliography{paper}
\bibliographystyle{iclr2018_conference}

\end{document}